\documentclass{article}

\usepackage[english]{babel}

\usepackage[letterpaper,top=2cm,bottom=2cm,left=3cm,right=3cm,marginparwidth=1.75cm]{geometry}

\usepackage{physics}
\usepackage{graphicx}
\usepackage{amsmath}
\usepackage[version=4]{mhchem}
\usepackage{longtable,tabularx}
\usepackage{bm}
\usepackage[ruled, vlined]{algorithm2e}
\usepackage{amssymb}
\usepackage{physics}
\usepackage{xcolor}

\newcommand{\Nhind}{{n}}
\newcommand{\Nfore}{{m}}
\newcommand{\Nx}{{p}}
\newcommand{\Nreal}{{k}}
\newcommand{\Nmodes}{{r}}
\newcommand{\eqn}[1]{equation (\ref{#1})}

\usepackage{hyperref}
\hypersetup{
    colorlinks=true,
    linkcolor=black,
    filecolor=magenta,      
    urlcolor=blue,
}
\usepackage{epstopdf}

\title{Data-Driven Forecasting of High-Dimensional Transient and Stationary Processes via Space–Time Projection}
\author{
	Oliver T. Schmidt\\ \small Department of Mechanical and Aerospace Engineering\\ \small University of California San Diego, La Jolla, CA 92093, USA}
\date{}

\begin{document}
\maketitle

\begin{abstract}
Space-Time Projection (STP) is introduced as a data-driven forecasting approach for high-dimensional and time-resolved data. The method computes extended space-time proper orthogonal modes from training data spanning a prediction horizon comprising both hindcast and forecast intervals. Forecasts are then generated by projecting the hindcast portion of these modes onto new data, simultaneously leveraging their orthogonality and optimal correlation with the forecast extension. Rooted in Proper Orthogonal Decomposition (POD) theory, dimensionality reduction and time-delay embedding are intrinsic to the approach. For a given ensemble and fixed prediction horizon, the only tunable parameter is the truncation rank---no additional hyperparameters are required. The hindcast accuracy serves as a reliable indicator for short-term forecast accuracy and establishes a lower bound on forecast errors. The efficacy of the method is demonstrated using two datasets: transient, highly anisotropic simulations of supernova explosions in a turbulent interstellar medium, and experimental velocity fields of a turbulent high-subsonic engineering flow. In a comparative study with standard Long Short-Term Memory (LSTM) neural networks---acknowledging that alternative architectures or training strategies may yield different outcomes---the method consistently provided more accurate forecasts. Considering its simplicity and robust performance, STP offers an interpretable and competitive benchmark for forecasting high-dimensional transient and chaotic processes, relying purely on spatiotemporal correlation information.
\end{abstract}

\section{Introduction}

Forecasting the evolution of high-dimensional real-world processes is a formidable challenge in a variety of fields ranging from atmospheric science to engineering and astrophysics. Such systems frequently exhibit nonlinear dynamics and chaos, imposing fundamental limits on predictability. When available and computationally tractable, physics-based numerical models---typically in the form of discretized systems of partial differential equations representing conservation laws for mass, momentum, and energy---offer powerful predictive capabilities. These numerical models are especially effective when combined with data assimilation techniques \cite{evensen1994sequential}, such as Kalman filtering \cite{kalmannew}, 4D-Var \cite{ledimet1986variational}, and ensemble-based methods \cite{toth1993ensemble}, to incorporate measurement data. The arguably most notable example is numerical weather prediction \cite{bauer2015quiet}.

This work addresses the complementary challenge of forecasting large spatiotemporal datasets through purely data-driven methods, regardless of their origin or the availability of governing physical models. This class of problems is often referred to as multivariate or multichannel time-series forecasting \cite{lutkepohl2005new, box2015time}. Widely known methods for forecasting such data include Autoregressive Integrated Moving Average (ARIMA), Vector Autoregression (VAR), Kalman filtering, state-space models \cite{durbin2012time}, and various extensions of Singular Spectrum Analysis (SSA;  \cite{golyandina2001analysis}). The choice of an appropriate forecasting method typically depends on the statistical nature of the data; transient, stationary, and cyclo-stationary processes are ubiquitous across many fields, each requiring different treatments or preprocessing techniques.

Many of these forecasting methods rely fundamentally on matrix decompositions, most commonly the Singular Value Decomposition (SVD) applied directly to data, Hankel or correlation matrices. As a result, these methods often directly make use of Proper Orthogonal Decomposition (POD), or indirectly leverage the same mathematical properties of optimality and orthogonality provided by SVD. While the original formulation of POD \cite{Lumley:1967,Lumley:1970} defines modes that are dependent on both space and time, the most commonly used variant, referred to simply as POD, separates the flow data into spatial modes and time-dependent expansion coefficients \cite{sirovich1987turbulence,aubry1991hidden}. To distinguish clearly between these different formulations, we refer to this widely used version as space-only or method-of-snapshots POD. Although the theoretical foundations of the most general form of space–time POD have long been established, algorithmic implementations and practical applications remain extremely rare, with \cite{gordeyev2013temporal,schmidt2019conditional} being exceptions. Only recently, the equivalence between space-time POD and the SVD of a Hankel matrix was explicitly demonstrated by \cite{Frame2023SpaceTimePOD}, who also showed that space-only POD and Spectral POD (SPOD; \cite{towne2018spectral,schmidt2020guide}) arise naturally as the limiting cases of space-time POD in the 
short- and long-time limits, respectively. Extended POD \cite{maurel2001extended}, a key enabler of this work, was initially proposed as a method to analyze correlated events in turbulent flows. It was first applied to investigate correlations between spatially distinct regions, and subsequently generalized and rigorously formalized by \cite{boree2003extended}. Similar to POD, Dynamic Mode Decomposition (DMD; \cite{schmid2010dynamic,rowley2009spectral}) is fundamentally a modal decomposition technique \cite{taira2017modal}. However, grounded in Koopman theory \cite{mezic2013analysis}, DMD yields a discrete propagation operator for the system state along with modes characterized by complex frequencies. This property enables direct forecasting by propagating the modes forward in time using their intrinsic oscillation frequencies and corresponding growth or decay rates. Similarly, SPOD yields modes associated with distinct frequencies, making them inherently suitable for forecasting without further modeling. That said, without additional modeling or rank truncation, SPOD forecasts reduce to repetitions of previously observed data due to the inherent stationarity assumption. A stochastic model combining SPOD for dimensionality reduction and Koopman theory for forecasting was recently proposed by \cite{chu2025stochasticreducedorderkoopmanmodel}. 

Machine learning (ML) methods represent another prominent category of forecasting techniques. Models specifically designed or readily applicable to time series forecasting include Recurrent Neural Networks (RNN; \cite{mandic2001recurrent}), Long Short-Term Memory networks (LSTM; \cite{hochreiter1997long}), and the simplified Gated Recurrent Units (GRU; \cite{cho2014learning}). Additional ML-based approaches include Temporal Convolutional Networks (TCN; \cite{lea2016temporal}), transformer-based architectures \cite{vaswani2017attention}, and Reservoir Computing (RC; \cite{jaeger2004harnessing,doan2021short}). Two contributions explicitly targeting high-dimensional physical system forecasting are \cite{vlachas2018data}, using LSTM, and \cite{Vlachas2020}, demonstrating RC. While these studies successfully forecast chaotic model systems such as the Lorenz system and the Kuramoto–Sivashinsky equation, they typically cannot be applied directly to large datasets arising from high-fidelity simulations or experimental measurements, such as Particle Image Velocimetry (PIV; \cite{raffel2007particle}). For such data, dimensionality reduction is usually required first, making ML-approaches sequential. One example, used later in this paper for comparison with the proposed STP method, incorporates POD for dimensionality reduction, followed by ML-based forecasting of the resulting POD coefficients representing latent temporal dynamics. This approach was explored in \cite{larioetal_2022_jcp} for both POD and SPOD.

\begin{figure}[ht!]
 \centering
 \includegraphics[width=1\textwidth]{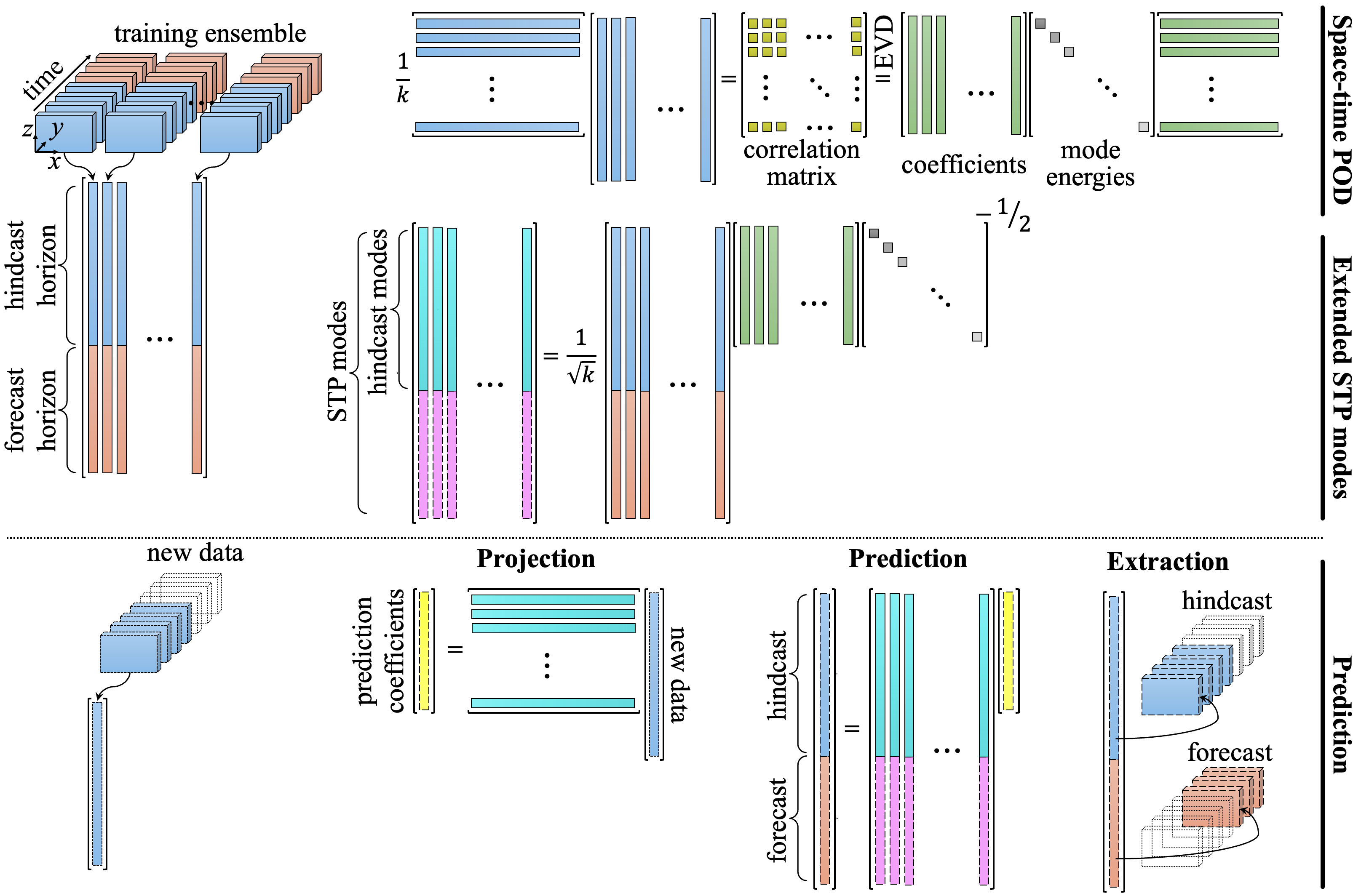}
 \caption{Schematic of the STP algorithm. Step 1: Construct the hindcast data matrix and solve the ensemble space-time POD eigenvalue problem, \eqn{eqn:evp}, to obtain the hindcast expansion coefficients. Step 2: Compute the STP (extended hindcast) modes using \eqn{eqn:phipm}. Step 3: Approximate the hindcast expansion coefficients via projection, \eqn{eqn:ahind}, and expand the predicted trajectory using \eqn{eqn:qstarpm}.}
\label{fig:sketch}
\end{figure}

This paper introduces Space–Time Projection (STP) as a forecasting method that exclusively relies on correlation information to make forecasts of high-dimensional fields. We begin by presenting the method formulation in Section~\ref{sec:methods}, accompanied by a visual summary in Figure~\ref{fig:sketch}. In the results section, Section~\ref{sec:results}, we demonstrate the method’s capabilities using two large datasets, both exhibiting highly chaotic dynamics, but one numerical and transient, the other experimental and stationary. First, in Section~\ref{sec:nova}, we examine the transient, non-stationary example---a supernova explosion evolving within a turbulent interstellar medium, simulated by \cite{hirashima20233d}. Second, in Section~\ref{sec:cavity}, we examine a statistically stationary case using time-resolved velocity field measurements from a Mach 0.6 turbulent cavity flow experiment by \cite{zhang2020spectral}, a canonical flow configuration representative of landing gear bays and cavities formed by high-lift devices on aircraft \cite{rowley2006dynamics}. In Section~\ref{sec:performance}, two important aspects of STP’s forecasting performance are evaluated: its sensitivity of forecast accuracy to the available training data size, and its forecast accuracy in direct comparison to LSTM regression neural networks trained on POD-reduced coefficients. Finally, in Section~\ref{sec:discussion}, we summarize the proposed approach, comment on the mathematical versus physical interpretation of the modes that form the basis, and outline future directions and potential applications.

\section{Methods}\label{sec:methods}

Given $\Nhind$ snapshots, $\vb{u}_1,\vb{u}_2,\dots,\vb{u}_\Nhind$, spanning the hindcast horizon, our goal is to predict the next $\Nfore$ snapshots, $\vb{u}_{\Nhind+1},\vb{u}_{\Nhind+2},\dots,\vb{u}_{\Nhind+\Nfore}$, over the forecast horizon. We refer to the combined hindcast and forecast horizons as the prediction horizon. The proposed method is based on an ensemble of $\Nreal$ realizations of the transient stochastic process we wish to predict, with each realization sampled over the prediction horizon. We denote the snapshot representing the system state at time $t_i$ ($i=1,2,\dots,\Nhind+\Nfore$) in the $j$-th realization ($j=1,2,\dots,\Nreal$) as
\begin{equation}
    \vb{u}_i^{[j]}=\vb{u}^{[j]}(\vb{x},t_i),
\end{equation} 
stored as a flattened vector comprising $\Nx$ degrees of freedom (typically, the total number of grid points multiplied by the number of variables). The entire sequence of $\Nhind+\Nfore$ snapshots for the $j$-th realization constitutes an episode, describing a single trajectory of the stochastic process. We assume that the ensemble mean,
\begin{equation}    \label{eqn:var_mean}
    \bar{\vb{u}}_{i} = \tfrac{1}{\Nreal}\sum_{j=1}^\Nreal \vb{u}_{i}^{[j]}
\end{equation}
computed over $\Nreal$ trajectories for each time instance $i$, is already subtracted from the data. This preprocessing step allows the eigenvalues to be interpreted as variances or, depending on the choice of variables and weighting, as mode energies. We construct the $\Nhind\Nx\times 1$ hindcast data vector $\vb{q}_- = [\vb{u}^T_1 \; \vb{u}^T_2 \cdots \vb{u}^T_\Nhind]^T$ by reshaping each snapshot $\vb{u}_i$ into a one-dimensional vector and then stacking them sequentially to form a single column vector. The subscript $(\cdot)_-$ denotes vectors and matrices spanning the hindcast horizon, $(\cdot)_+$ the forecast horizon, and $(\cdot)_\pm$ the entire prediction horizon, encapsulating both the hindcast and forecast. 

Given an ensemble of $\Nreal$ realizations of the stochastic process, we construct the $\Nhind\Nx\times\Nreal$ hindcast data matrix, 
\begin{eqnarray}
&
    \vb{Q}_- = \qty[ \vb{q}_-^{[1]} \; \vb{q}_-^{[2]} \; \cdots \; \vb{q}_-^{[\Nreal]}]
    = \mqty[
    \vb{u}^{[1]}_1 & \vb{u}^{[2]}_1 & \cdots & \vb{u}^{[\Nreal]}_1 \\ 
    \vb{u}^{[1]}_2 & \vb{u}^{[2]}_2 & \cdots & \vb{u}^{[\Nreal]}_2 \\ 
    \vdots & \vdots &  \ddots & \vdots \\
    \vb{u}^{[1]}_\Nhind & \vb{u}^{[2]}_\Nhind & \cdots & \vb{u}^{[\Nreal]}_\Nhind],
&
\end{eqnarray}
and the $(\Nhind+\Nfore)\Nx\times\Nreal$ prediction data matrix,
\begin{eqnarray}\label{eqn:Qpm}
&
    \vb{Q}_\pm
    = \mqty[ \vb{q}_-^{[1]} & \vb{q}_-^{[2]} & \cdots & \vb{q}_-^{[\Nreal]} \\ \vb{q}_+^{[1]} & \vb{q}_+^{[2]} & \cdots & \vb{q}_+^{[\Nreal]}]
    = \mqty[
    \vb{u}^{[1]}_1 & \vb{u}^{[2]}_1 & \cdots & \vb{u}^{[\Nreal]}_1 \\ 
    \vb{u}^{[1]}_2 & \vb{u}^{[2]}_2 & \cdots & \vb{u}^{[\Nreal]}_2 \\ 
    \vdots & \vdots &  \ddots & \vdots \\
    \vb{u}^{[1]}_\Nhind & \vb{u}^{[2]}_\Nhind & \cdots & \vb{u}^{[\Nreal]}_\Nhind \\
    \vb{u}^{[1]}_{\Nhind+1} & \vb{u}^{[2]}_{\Nhind+1} & \cdots & \vb{u}^{[\Nreal]}_{\Nhind+1} \\ 
    \vdots & \vdots &  \ddots & \vdots \\
    \vb{u}^{[1]}_{\Nhind+\Nfore} & \vb{u}^{[2]}_{\Nhind+\Nfore} & \cdots & \vb{u}^{[\Nreal]}_{\Nhind+\Nfore}],
&
\end{eqnarray}
which includes full episodes spanning both the hindcast and forecast horizons. Analogous to the hindcast data, the forecast data is flattened into the $\Nfore\Nx\times 1$ vector $\vb{q}_+ = [\vb{u}^T_{\Nhind+1} \; \vb{u}^T_{\Nhind+2} \cdots \vb{u}^T_{\Nhind+\Nfore}]^T$.

At the core of the method is the ensemble space-time POD of the hindcast data matrix $\vb{Q}_-$. Like other POD variants, the space-time POD problem can be solved using the eigenvalue decomposition of either the spatiotemporal or ensemble correlation matrix, or directly through the SVD of the data matrix. Since the ensemble size $\Nreal$ is typically much smaller than the total number of spatiotemporal degrees of freedom, $\Nhind\Nx$, we perform the computation using the Hermitian $\Nreal\times\Nreal$ ensemble sample correlation matrix,
\begin{equation}\label{eqn:chind}
    \vb{C}_- = \tfrac{1}{\Nreal}\vb{Q}_-^H\vb{W}\vb{Q}_-,
\end{equation}
and compute its eigenvalue decomposition,
\begin{equation}\label{eqn:evp}
    \vb{C}_-\vb{\Psi}_- = \vb{\Psi}_-\vb{\Lambda},
\end{equation}
to obtain the ensemble coefficient matrix, scaled such that $\vb{\Psi}_-^H\vb{\Psi}_-=\vb{I}$. For generality, the diagonal weight matrix $\vb{W}$ is introduced in equation (\ref{eqn:chind}) to account for spatial and variable weighting. The diagonal eigenvalue matrix $\vb{\Lambda}=\mathrm{diag}(\lambda_1,\lambda_1,\dots,\lambda_\Nreal)$ contains the mode energies, ranked by their magnitude. Note that, depending on the choice of variables, these eigenvalues may not strictly represent physical energy; rather, they quantify the captured variance of each mode. In any case, they indicate the relative importance of individual modes and can always be interpreted as the squared amplitudes of the modes. 

If rank reduction is desired—whether to suppress noise by discarding low-energy components or to reduce computational complexity—we can truncate the representation to retain only the leading $\Nmodes < \Nreal$ components. Specifically, we truncate the columns of the coefficient matrix $\vb{\Psi}_-$ and reduce the eigenvalue matrix to $\vb{\Lambda}=\mathrm{diag}(\lambda_1,\lambda_1,\dots,\lambda_\Nmodes)$. With this reduced representation, all subsequent steps---mode reconstruction, projection, and forecasting---are carried out using the truncated basis. We omit explicit notation for this rank truncation to keep the presentation concise. First, the \emph{hindcast modes} are obtained by expanding the hindcast data using the ensemble coefficients as
\begin{equation}\label{eqn:phihind}
    \vb{\Phi}_- = \tfrac{1}{\sqrt{\Nreal}}\vb{Q}_-\vb{\Psi}_-\vb{\Lambda}^{-\frac{1}{2}} \quad \text{(hindcast modes)},
\end{equation}
where the scaling ensures that the hindcast modes are orthonormal with $\vb{\Phi}_-^H \vb{W}\vb{\Phi}_- = \vb{I}$. Since the modal basis $\vb{\Phi}_-$ spans the same subspace as the data $\vb{Q}_-$, we can express the latter as a linear combination of the modes:
\begin{equation}\label{eqn:Qexp}
    \vb{Q}_- = \vb{\Phi}_- \vb{A}_-.
\end{equation}
The matrix of expansion coefficients $\vb{A}_-$ can be related to $\vb{\Psi}_-$ via $\vb{A}_-=\sqrt{\Nreal}\vb{\Lambda}^\frac{1}{2}\vb{\Psi}_-^H$, or, alternatively, obtained by projecting the data directly onto the modes as
\begin{eqnarray}\label{eqn:Aproj}
    \vb{A}_-=\vb{\Phi}_-^H \vb{W} \vb{Q}_-,
\end{eqnarray} 
and has the orthogonality property $\frac{1}{\Nreal}\vb{A}_-^H\vb{A}_-=\vb{\Lambda}$. Together with the orthonormality of the modes, this property makes clear that we can interpret the square root of the eigenvalues as mode amplitudes. 

To make predictions over the forecast horizon, we define the \emph{STP modes} as the extended hincast modes,
\begin{equation}\label{eqn:phipm}
    \vb{\Phi}_\pm^\star = \tfrac{1}{\sqrt{\Nreal}}\vb{Q}_\pm\vb{\Psi}_-\vb{\Lambda}^{-\frac{1}{2}} \quad \text{(STP modes)},
\end{equation}
obtained analogously to \eqn{eqn:phihind} by expanding the ensemble data of the full trajectories into extended modes, $\vb{\Phi}_\pm^\star$, which span both the hindcast and forecast horizons. This is done by replacing $\vb{Q}_-$ in \eqref{eqn:phihind} with $\vb{Q}_\pm$, while keeping the same hindcast expansion coefficients. We use the superscript $(\cdot)^\star$ to denote prediction quantities that are analogous to but deviate from their exact counterparts. Note that $\vb{\Phi}_\pm^\star$ are not the space-time modes $\vb{\Phi}_\pm$ related to $\vb{C}_\pm$, but rather extended hindcast modes. In the expansion \eqref{eqn:phipm}, the hindcast components remain unchanged, meaning the first $\Nhind\Nx$ elements of each STP mode $\boldsymbol{\phi}_\pm^\star$ (the columns of $\vb{\Phi}_\pm^\star$) exactly match the corresponding hindcast mode $\boldsymbol{\phi}_-$. However, these modes are extended over the forecast horizon by an additional $\Nfore\Nx\times1$ vector, $\boldsymbol{\phi}_+^\star$, which retains the portion---and only that---of the forecast data that is correlated with the hindcast. This key correlation property of extended POD was elegantly proven by \cite{boree2003extended} and directly applies to the extended space-time, or STP modes $\boldsymbol{\phi}_\pm^{\star} = [\boldsymbol{\phi}_-^T\;\boldsymbol{\phi}_+^{\star T}]^T$. We refer to Proposition 1 in \cite{boree2003extended} for a formal proof. 

Given a new hindcast trajectory, $\vb{{q}}_-^\textrm{new}$, consisting of $n$ stacked snapshots, our goal is to forecast the next $\Nfore$ time instances of the flow, $\vb{{q}}_+^\star$. To compute the forecast using an expansion in the STP mode basis, $\vb{\Phi}_\pm$, we need the corresponding expansion coefficients, $\vb{{a}}_\pm^\star$, which are unknown. The key idea is to approximate $\vb{{a}}_\pm^\star$ using $\vb{{a}}_-^\star$, the hindcast expansion coefficients, and leverage the correlation property of extended POD to ensure an optimal forecast. The hindcast expansion coefficients $\vb{{a}}_-^\star$ are obtained in accordance with \eqn{eqn:Aproj} by projecting the new hindcast data onto the precomputed hindcast basis:
\begin{equation}\label{eqn:ahind}
    \vb{{a}}_-^\star = \vb{\Phi}_-^H \vb{W} \vb{{q}}_-^\textrm{new}.
\end{equation}
Under the approximation that $\vb{{a}}_\pm^\star \approx \vb{{a}}_-^\star$, we can readily reconstruct the predicted full trajectory, $\vb{{q}}_\pm^\star \approx \vb{{q}}_\pm$, as
\begin{equation}\label{eqn:qstarpm}
    \vb{{q}}_\pm^\star = \vb{\Phi}_\pm^\star \vb{{a}}_-^\star,
\end{equation}
that is, by expanding the STP mode basis using the hindcast coefficients, see \eqn{eqn:Qexp}. The prediction, $\vb{{q}}_\pm^\star = [\vb{{q}}_-^{\star T} \; \vb{{q}}_+^{\star T}]^T$, can then be separated into the hindcast, $\vb{{q}}_-^{\star}$, and the forecast, $\vb{{q}}_+^\star$. Alternatively, we can separate equation (\ref{eqn:qstarpm}) and compute the hindcast and forecast components separately as $\vb{{q}}_-^\star = \vb{\Phi}_-^\star \vb{{a}}_-^\star$ and $\vb{{q}}_+^\star = \vb{\Phi}_+^\star \vb{{a}}_-^\star$, respectively. A difference between the new data, $\vb{{q}}_-^\textrm{new}$, and the hindcast, $\vb{{q}}_-^{\star}$, is expected for real-world stochastic physical processes, where the new data may not lie within the span of the training data used to compute the STP basis via equation \eqref{eqn:evp}. As the examples will demonstrate, this hindcast error provides valuable insight into the accuracy of the forecast. The key steps of the algorithm are visually summarized in Figure \ref{fig:sketch}.  For simplicity, uniform weighting with $\vb{W}=\vb{I}$ is assumed for Figure \ref{fig:sketch} and the definitions of following error metrics.

We use the root mean square error (RMSE) to quantify the difference between the true state, $\vb{{u}}_{i} = \vb{{u}}(\vb{x},t_i)$, and the predicted fields, $\vb{{u}}_{i}^\star = \vb{{u}}^\star(\vb{x},t_i)$, at the $i$-th time instant. The RMSE for a single trajectory is hence defined as
\begin{equation}\label{eqn:rms}
    e_{\text{RMS},i} = \tfrac{1}{\sqrt{\Nx}}\|\vb{{u}}_{i} - \vb{{u}}_{i}^\star \|_2^2, 
\end{equation}   
the mean RMSE over an ensemble of $k$ trajectories by
\begin{equation}    \label{eqn:rms_mean}
    \bar{e}_{\text{RMS},i}= \tfrac{1}{\Nreal}\sum_{j=1}^\Nreal e_{\text{RMS},i}^{[j]},
\end{equation}
and the corresponding standard deviation, which quantifies the spread of the error, is calculated as
\begin{equation}    \label{eqn:rms_std}
\sigma_{e_{\mathrm{RMS},i}} = \sqrt{\tfrac{1}{k - 1} \sum_{j=1}^k 
(e_{\mathrm{RMS},i}^{[j]} - \bar{e}_{\mathrm{RMS},i})^2}.
\end{equation}
Together, these metrics provide a comprehensive assessment of the prediction accuracy, both in terms of average performance and variability across trajectories.

\section{Results}\label{sec:results}

\begin{figure}[ht!]
 \centering
\includegraphics[width=1\textwidth,trim={0 11cm 0 5cm},clip]{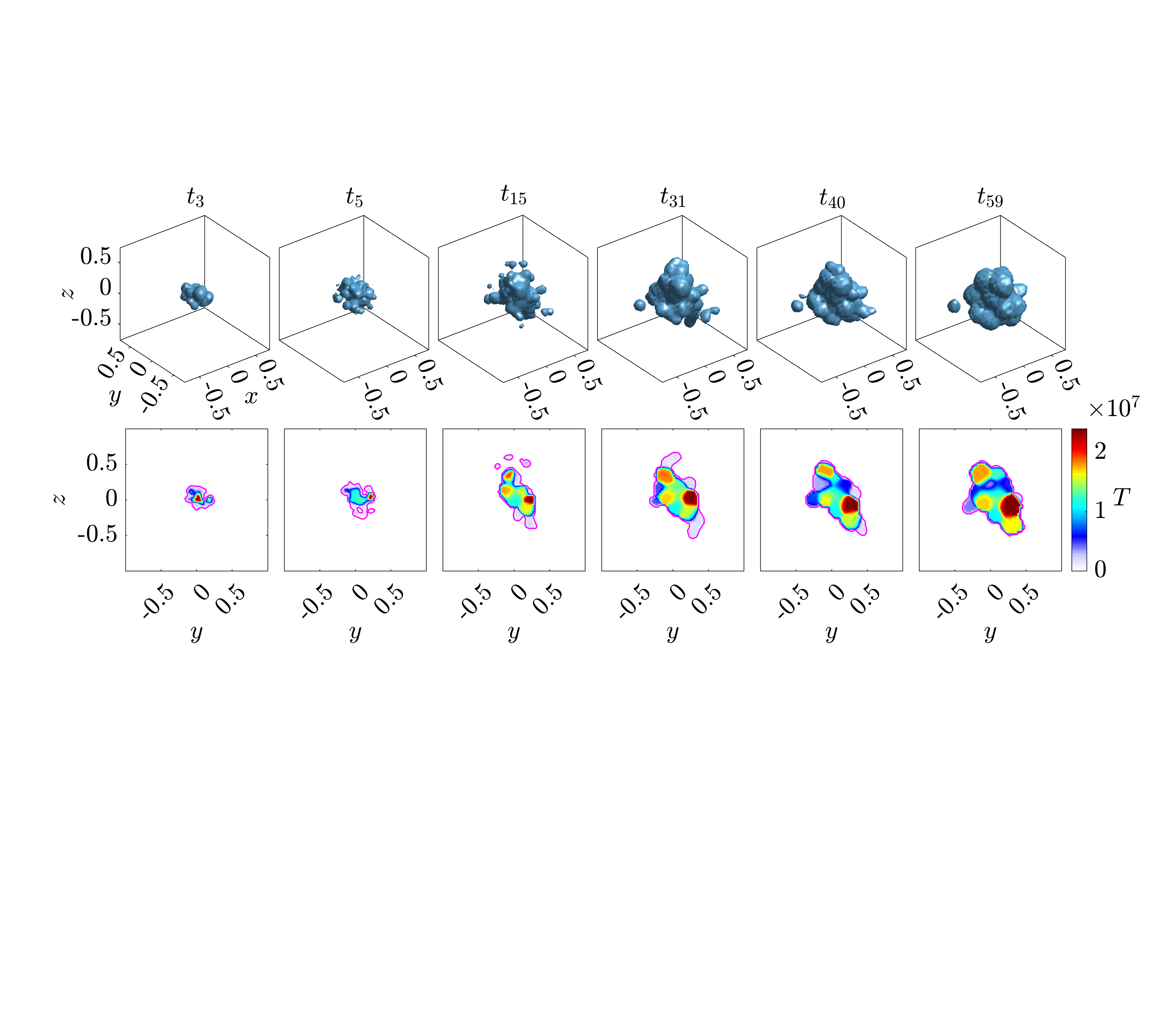}
 \caption{Overview of the numerical supernova simulation: The top row shows instantaneous isocontours of temperature at 1\% of its maximum value, highlighting the expanding supernova shell at six representative time instances. The bottom row presents the corresponding temperature fields in the $y$-$z$ plane at $x=0$.  The magenta contour line indicates the isovalue used in the 3D visualization of the shell. The first realization from the training dataset is shown as an example.}
\label{fig:supernova_flow_overview}
\end{figure}

\begin{figure}[ht!]
 \centering
 \includegraphics[width=1\textwidth]{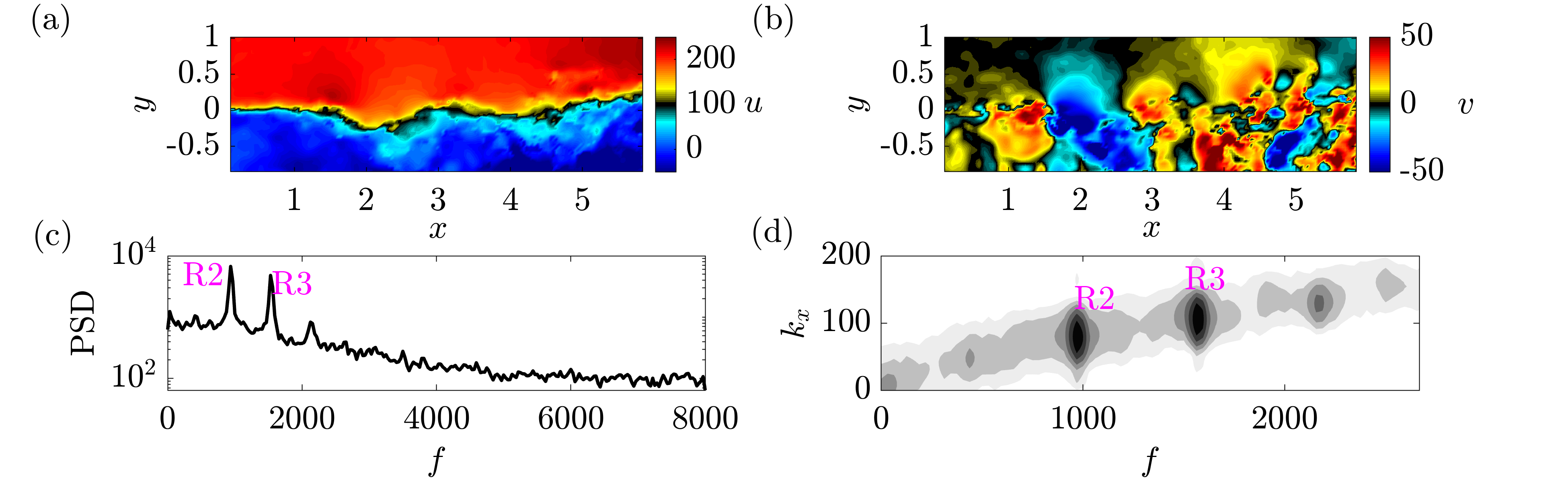}
 \caption{Overview of experimental cavity flow: (a) instantaneous streamwise velocity $u$; (b) instantaneous normal velocity $v$; (c) power spectrum at $(x,y)=(5,0)$; (d) frequency–wavenumber diagram along $(x,y=0)$. The peaks identified in (c) and (d) correspond to the two dominant Rossiter tones. The incoming flow travels over the cavity ($y>0$) at Mach 0.6 and recirculates within the cavity walls ($y<0$) while undergoing violent oscillations.}
\label{fig:flow}
\end{figure}

STP is demonstrated on two large, distinct datasets, each presenting unique challenges. The first example, discussed in Section~\ref{sec:nova}, involves numerical data of a supernova explosion \cite{hirashima20233d} with a highly anisotropic shell rapidly expanding into a turbulent interstellar medium, whose localized and transient features make it particularly challenging. The second example, examined in Section~\ref{sec:cavity}, considers experimental velocity field measurements of the flow over and inside an open cavity at high speeds \cite{zhang2020spectral}, where the main difficulties arise from the broadband nature of the stationary turbulent flow and the convective behavior of its coherent structures. Figures \ref{fig:supernova_flow_overview} and \ref{fig:flow} introduce the two cases and highlight these challenges. Since both datasets are sampled on equidistant grids and we are only concerned with relative comparisons, we use uniform weighting with $\vb{W}=\vb{I}$.

\subsection{Supernova Explosion}\label{sec:nova}
The first example considers a supernova explosion in a turbulent interstellar medium, where a blast wave propagates through a dense, cool gas cloud, generating complex and anisotropic shell structures. The simulations, conducted by \cite{hirashima20233d} using the Smoothed Particle Hydrodynamics (SPH) code ASURA-FDPS \cite{saitoh2008toward,iwasawa2016implementation}, were obtained from the The Well data repository \cite{ohana2024thewell}. The setup models a supernova—the explosion of a massive star—within a compressible monatomic ideal gas, governed by the equation of state with a specific heat ratio $\gamma=5/3$. The simulated gas represents the interstellar medium of the Milky Way galaxy with an initial gas-particle mass of one solar mass. To initiate the explosion, a thermal energy of $10^{51}$ erg is deposited at the domain center, rapidly heating the gas to temperatures around $10^7$ K and generating a blast wave. The initial conditions feature 820 randomly seeded molecular clouds with turbulent statistics following Burgers turbulence. 

For this demonstration, we use the temperature field $T(\vb{x},t)$ as the sole variable. For each snapshot, the ensemble mean, calculated using \eqn{eqn:var_mean}, is subtracted and we define the state vector for the $i$-th snapshot as
\[\vb{u}_i=T'_i,\]
where $T'_i$ is the flattened vector of the fluctuating temperature. The repository provides temperature data at 59 time steps on a uniform Cartesian $64\times64\times64$ grid, spanning 60 pc in each direction. For this single scalar variable, temperature, each snapshot hence contains $\Nx=64^3$ degrees of freedom. Although the time step size varies between 100 and 10,000 years, this variability does not affect the STP prediction as long as time steps remain consistent across episodes. We use 400 trajectories from the training set (with an initial gas-particle mass of one solar mass) to construct the STP mode basis, and the 50 trajectories from the test set for error analysis. The trajectory visualized in Figure \ref{fig:supernova_flow_overview} at five representative time instances illustrates the evolution of the complex shell structure during the rapid expansion. 

\begin{figure}[ht!]
 \centering
\includegraphics[width=1\textwidth,trim={0 0 0 0},clip]{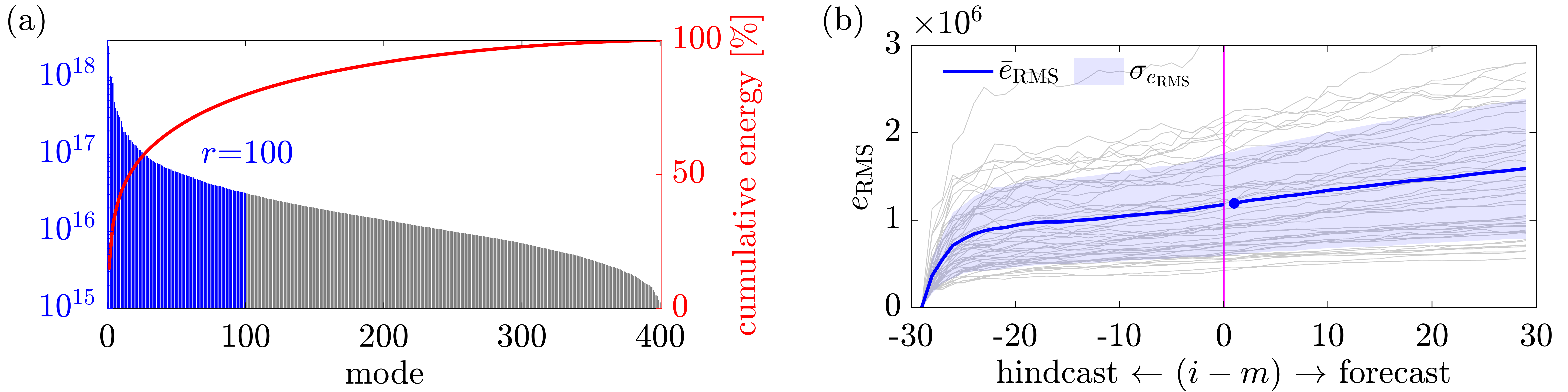}
 \caption{STP prediction of the supernova simulation with a hindcast horizon of $\Nhind=30$ and forecast horizon of $\Nfore=29$: (a) hindcast mode variance; (b) prediction error, where gray lines represent individual forecasts, the blue line indicates the mean, the solid circle indicates the first time step of the forecastand, and the light blue band shows the standard deviation. The leading $r=100$ retained modes, out of a total of 400, capture roughly 80\% of the total variance. In (b), the time step index $i$ is used because the time steps are non-uniform.}
\label{fig:supernova_spectrum_and_error}
\end{figure}
\begin{figure}[ht!]
 \centering
\includegraphics[width=1\textwidth,trim={0 1cm 0 0},clip]{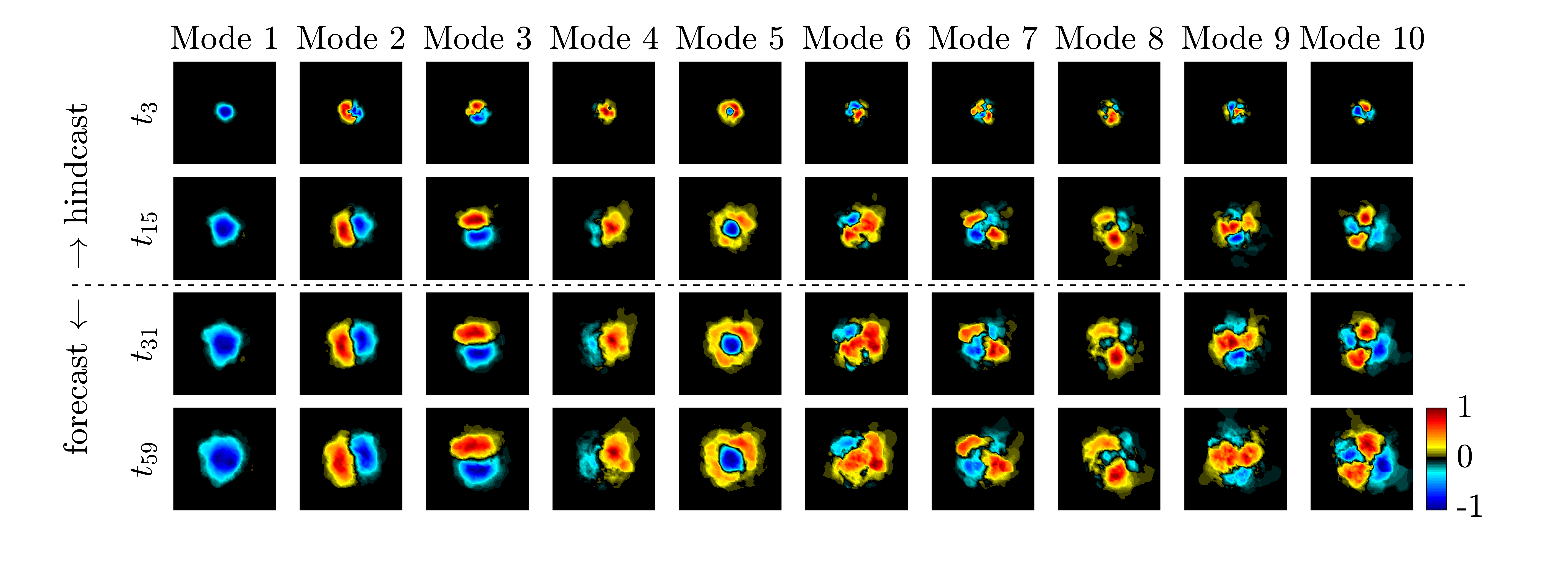}
 \caption{STP modes $\vb{\Phi}_\pm^\star$ of the supernova data for a hindcast horizon of $\Nhind=30$ and a forecast horizon of $\Nfore=29$. A prediction using this basis is shown in Figure \ref{fig:supernova_prediction}. The domain is the same as in Figure \ref{fig:supernova_flow_overview}, with $(y,z) \in [-1,1]\times[-1,1]$ at $x=0$. The fluctuating temperature fields are normalized by their maximum absolute values.}
\label{fig:supernova_modes}
\end{figure}

The supernova explosion is a transient phenomenon characterized by large-scale structural changes and non-repeating dynamics. Each trajectory captures distinct, nontrivial, anisotropic, and evolving shell formations. Our standard forecasting task for such transient data involves predicting the remaining $\Nfore$ time instances of the forecast horizon, given an observed hindcast horizon of $\Nhind$ time instances. For the first demonstration, we set the hindcast interval to $\Nhind=30$ and predict the remaining $\Nfore = 29$ snapshots. Figure \ref{fig:supernova_spectrum_and_error}(a) shows the hindcast eigenvalue spectrum computed according to \eqn{eqn:evp}. The most correlated structure captured by the leading mode corresponds to a correction to the ensemble mean, as later discussed in the context of Figure \ref{fig:supernova_modes}. After a sharp initial drop in mode variance over the first few modes, the eigenvalues exhibit a near-algebraic decay over a large portion of the spectrum. The first $\Nmodes=100$ modes capture approximately 80\% of the total variance and 344 are needed to reach 95\%, as can be seen in Figure \ref{fig:supernova_spectrum_and_error}(a).

Figure \ref{fig:supernova_spectrum_and_error}(b) shows the RMS errors for the 50 test trajectories, as well as their mean, defined by \eqn{eqn:rms_mean}, for the forecasted temperature fields. The standard deviation of the error indicates the spread in forecast accuracy across different trajectories. The error is plotted over the time step index relative to the start of the forecast horizon, which the magenta vertical line separates from the hindcast horizon. Initially, the error increases rapidly over approximately the first 10 time instances of the hindcast, then continues to grow more slowly. Notably, there is no sudden change at the transition from hindcast to forecast, suggesting that the hindcast error is a good indicator of the forecast error.

\begin{figure}[ht!]
 \centering
\includegraphics[width=1\textwidth,trim={0 0 0 0},clip]{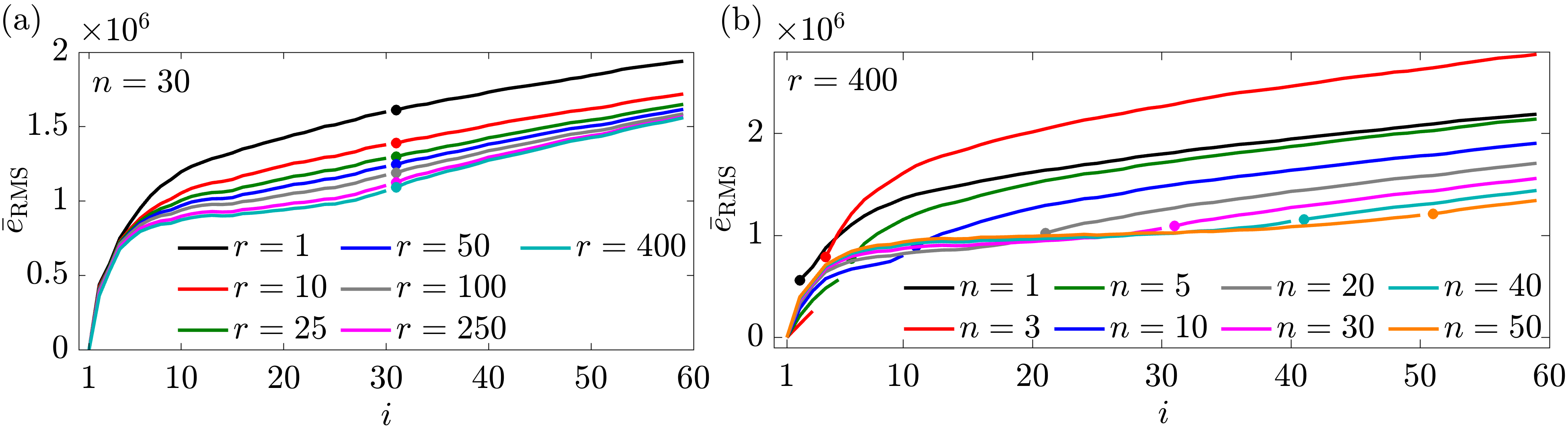}
 \caption
 {Study of the prediction error in the supernova prediction: (a) mean prediction error as a function of hindcast interval length $\Nhind$ without basis truncation; (b) mean prediction error for different numbers $r$ of retained modes, with a fixed hindcast interval length of $\Nhind=30$. In all cases, the entire trajectory is considered, such that $\Nhind+\Nfore = 59$. Solid circles indicate the first time step of the forecast.}
\label{fig:supernova_nhind_and_r_Study}
\end{figure}
For a fixed hindcast horizon, the only parameter in this method is the rank 
$\Nmodes$ of the STP basis, that is, the number of retained STP modes. Figure \ref{fig:supernova_nhind_and_r_Study}(a) shows the mean forecast error, computed from 50 test episodes, for varying $\Nmodes$ at a fixed 
$\Nhind=30$. In this dataset—unlike the second example—the smallest error is achieved using the full rank of 
$\Nmodes=400$. However, a substantial reduction to 250 modes only slightly increases the error. By contrast, the rank-1 reconstruction yields the largest error, though it does not differ substantially from higher-rank forecasts as the ensemble mean already represetns the most energetic feature. Figure \ref{fig:supernova_nhind_and_r_Study}(b) shows that the forecast error for varying hindcast lengths largely follows the expected trend: longer hindcast horizons lead to more accurate forecasts. The only exception is when $\Nhind=3$, which results in a significantly larger forecast error.

Figure \ref{fig:supernova_modes} presents the first ten STP modes, $\vb{\Phi}_\pm^\star$, out of a total of 400, at four representative time instances—two from the hindcast horizon and two from the forecast horizon. The ensemble mean, which is subtracted from the data, represents the average rapid expansion of the supernova burst and consequently accounts for the largest fraction of the variance. The first mode similarly exhibits an expanding spherical pattern, providing a correction to the ensemble mean. The remaining modes build a basis that captures the variability of individual trajectories about the mean, forming a hierarchical structure of multipoles with increasing geometric complexity. Higher modes also tend to be noisier, suggesting reduced convergence.
 
\begin{figure}[ht!]
 \centering
\includegraphics[width=1\textwidth,trim={0 1.5cm 0 0},clip]{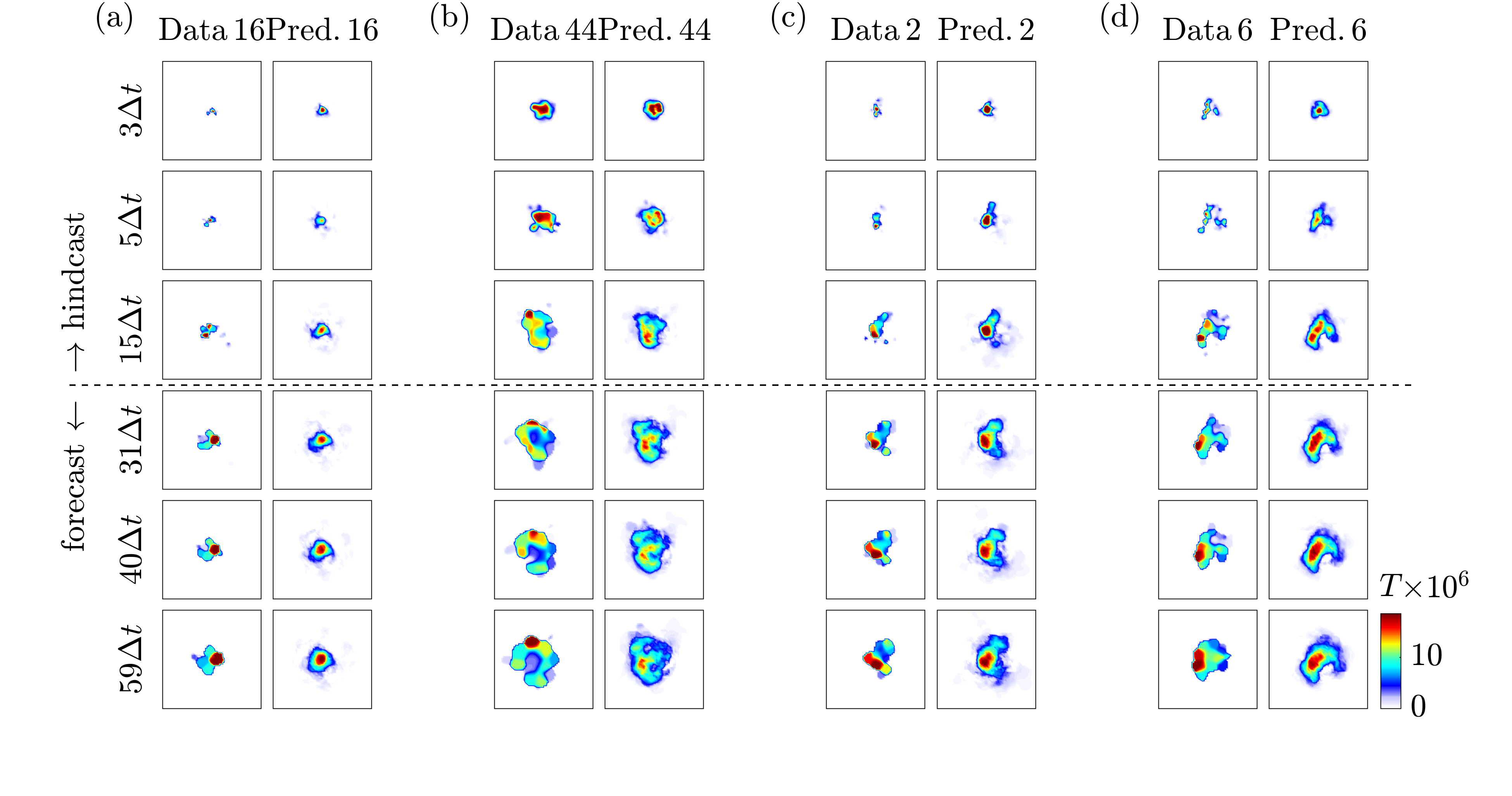}
 \caption{
 Prediction of the supernova using a hindcast horizon of $\Nhind=30$ and a forecast horizon of $\Nfore=29$: (a) trajectory with the lowest mean prediction error; (b) trajectory with the highest mean prediction error; (c, d)  trajectories with intermediate errors, selected to illustrate variability. The visualization domain is the same mid-cut plane as in Figure \ref{fig:supernova_flow_overview}, with $(y,z) \in [-1,1]\times[-1,1]$ at $x=0$.}
\label{fig:supernova_prediction}
\end{figure}

Figure \ref{fig:supernova_prediction} displays four representative supernova prediction trajectories, obtained using a hindcast horizon of $\Nhind=30$, a forecast horizon of $\Nfore=29$, and a full-rank STP basis with $\Nmodes=\Nreal=400$. Specifically, we selected trajectories corresponding to the lowest and highest forecast errors at the end of the forecast horizon, as well as two representative episodes that exhibit distinct features. A comparison between the ground truth, i.e., the test data, and the predictions reveals that the primary topological features of the individual explosions are well captured across both the hindcast and forecast horizons. However, certain localized features—such as the nearly perfect circular spots observed in the test data associated with growth from localized nuclei—appear blurred in the predictions. This is expected, given that such localized structures are challenging to represent accurately with a modal basis derived from very limited training data exhibiting substantially different characteristics. Considering these limitations, the overall evolution of these diverse trajectories is, arguably, captured remarkably well.


\subsection{Open Cavity Flow}\label{sec:cavity}
The second example is a time-resolved PIV (TR-PIV) experiment performed by \cite{zhang2020spectral} to obtain the two-dimensional velocity field in the center plane of a Mach 0.6 turbulent flow over an open cavity with length-to-depth and width-to-depth ratios of $L/D = 6$ and $W/D = 3.85$, respectively. We denote the streamwise and wall-normal velocities by $u(x,y,t)$ and $v(x,y,t)$, respectively, and define the state for the $i$-th snapshot as
\[\vb{u}_i = \mqty[u'_i \\ v'_i],\]
where the prime indicates the mean-subtracted (fluctuating) component of each velocity component, flattened and stacked into a single vector. Since the flow field is stationary, the ensemble and temporal means are equivalent, and we can subtract the latter. The database comprises 16,000 flow fields acquired at 16 kHz in a field of view resolved by a $160\times57$ grid. To eliminate regions with missing data, we removed three points from each lateral boundary, resulting in snapshots with $\Nx=18240$ degrees of freedom each. Further details on the experimental setup and hardware are reported in \cite{zhang2020spectral}. 

This flow differs fundamentally from the first example because it is statistically stationary rather than transient, and in that it is inherently noisy. Figure \ref{fig:flow} illustrates these typical characteristics of a turbulent stationary flow. The instantaneous velocity fields shown in \ref{fig:flow}(a,b) appear chaotic, and the power spectrum in \ref{fig:flow}(b) computed at $(x,y)=(5,0)$ is broadband with several distinct tones emerging from the turbulent background. The two prominent peaks labeled R2 and R3 at frequencies $f_\textrm{R1}=945$ and $f_\textrm{R2}=1547$ correspond to the second and third Rossiter tones, stemming from a resonance between downstream-traveling shear-layer instability waves and upstream-traveling acoustic waves \cite{rossiter1964wind}. The frequency-wavenumber diagram in \ref{fig:flow}(c) further enables approximation of the wavenumbers, $k_{x,\mathrm{R1}}=74$ and $k_{x,\mathrm{R2}}=80$, and thus the phase velocities, $c_{\mathit{ph},\mathrm{R1}}=74$ and $c_{\mathit{ph},\mathrm{R2}}=98$, of these vortical structures. From these values, we determine that the waves require about $32\Delta t$ and $26\Delta t$ to traverse the cavity.

When predicting a stationary flow---where statistics remain constant over time---we have more flexibility in choosing a hindcast horizon. However, the presence of large-scale coherent structures, such as the two dominant Rossiter modes, helps guide this choice and sets expectations for a realistic forecast horizon. Note that the forecast accuracy is independent of the forecast horizon.

\begin{figure}[ht!]
 \centering
 \includegraphics[width=1\textwidth]{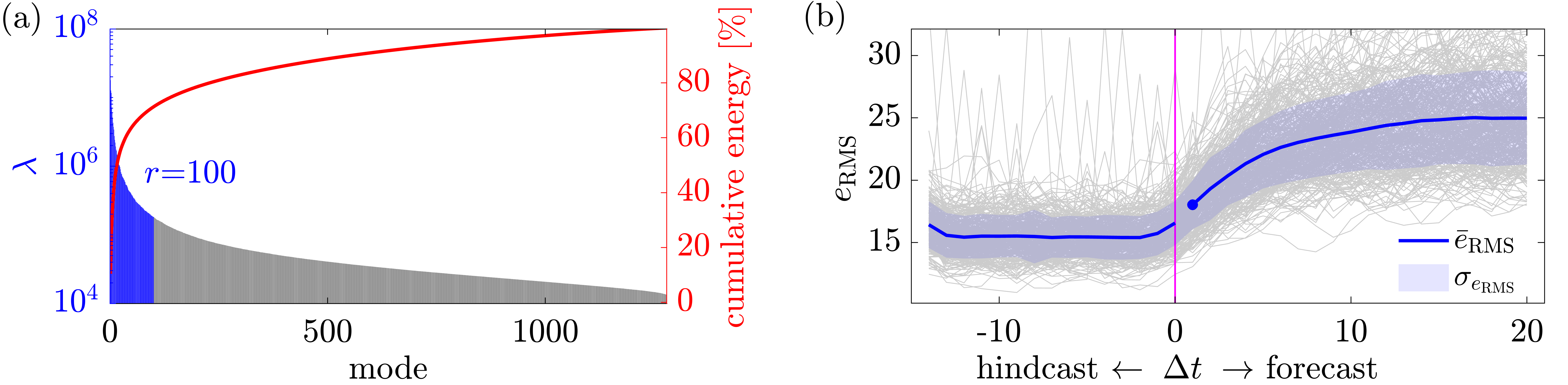}
 \caption{STP prediction of the cavity flow with a hindcast and forecast horizon of $\Nhind=15$ and $\Nfore=20$: (a) hindcast mode energy; (b) prediction error, where gray lines represent the 316 individual forecasts, the blue line indicates their mean, and the light blue band their standard deviation. The leading $\Nmodes=100$ retained modes, out of a total of 1278, capture 71\% of the total energy.}
\label{fig:spectrum_and_error}
\end{figure}
Before investigating the dependency of the forecast error on $\Nhind$ and $\Nmodes$ in more detail, we begin with a hindcast horizon of $\Nhind=15$, a forecast horizon of $\Nfore=20$, and retain the leading $\Nmodes=100$ modes as a baseline. The data is split in the standard 80\%/20\% manner, and overlapping segments are employed to obtain a sufficiently large number of training and testing episodes---a common practice for stationary data borrowed from spectral estimation \cite{welch1967use}. Specifically, each new episode is initiated ten time steps after the start of the previous episode, resulting in a training ensemble of size $\Nreal=1278$ and 316 testing episodes for independent forecast error quantification. Figure \ref{fig:spectrum_and_error}(a) presents the hindcast energy spectrum. Since the state comprises the fluctuating velocity components in the PIV plane, the eigenvalues directly capture the resolved turbulent kinetic energy. The mean prediction error, shown in Figure \ref{fig:spectrum_and_error}(b), exhibits distinct behavior for the hindcast and forecast segments: the hindcast error remains nearly constant, with minor boundary effects, while the forecast error monotonically increases for about ten time steps before settling at a constant level. At this point, the variance of individual trajectories is not accurately captured and the predictive capacity is reached.

\begin{figure}[ht!]
 \centering
 \includegraphics[width=1\textwidth,trim={0 0.5cm 0 0},clip]{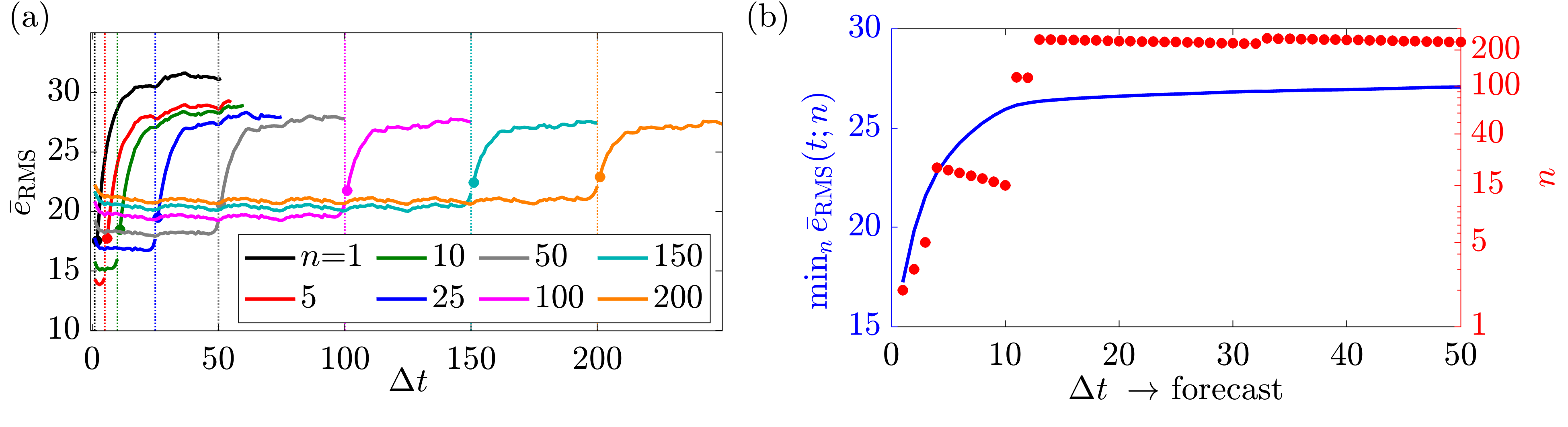}
 \caption{Study of the influence of hindcast interval length $n$ on prediction error for the cavity flow: (a) mean prediction error; (b) minimum mean error as a function of forecast time and corresponding optimal hindcast length. The forecast horizon is set to $\Nfore=20$.}
\label{fig:error_25skip_1to200}
\end{figure}
Figure \ref{fig:error_25skip_1to200} summarizes a parameter study on the influence of the hindcast horizon length on the prediction error. Panel \ref{fig:error_25skip_1to200}(a) displays the mean RMS error for eight different hindcast horizons ranging from $\Nhind=1$ to 200. The variability in the mean error arises from the increased sampling interval, as a new episode is initiated every 25 time steps to keep computational costs manageable. Two qualitative trends are immediately evident: the hindcast error increases with increasing $\Nhind$, while the forecast error decreases until it plateaus for $\Nhind > 10$. This suggests that forecast quality for a given forecast time depends on the hindcast horizon length, motivating the more detailed analysis in Figure \ref{fig:error_25skip_1to200}(b). Here, the minimum mean RMS error over the range $1\leq\Nhind\leq250$ is shown, with the optimal $\Nhind$ for each forecast length indicated by red filled circles. The results confirm that very short hindcast horizons are preferable for very short forecast times. In practice, a hindcast horizon of $\Nhind=15$ represents a good compromise between short- and long-term forecast accuracy, especially considering that forecasts beyond approximately $15\Delta t$ are intrinsically challenging given the inherent time scales involved. We hypothesize that the observed correspondence between the optimal hindcast horizon and the achievable forecast horizon is determined by the time scale over which the system exhibits strong spatiotemporal correlations, with the characteristic convection time establishing an upper limit.

\begin{figure}[ht!]
 \centering
 \includegraphics[width=1\textwidth]{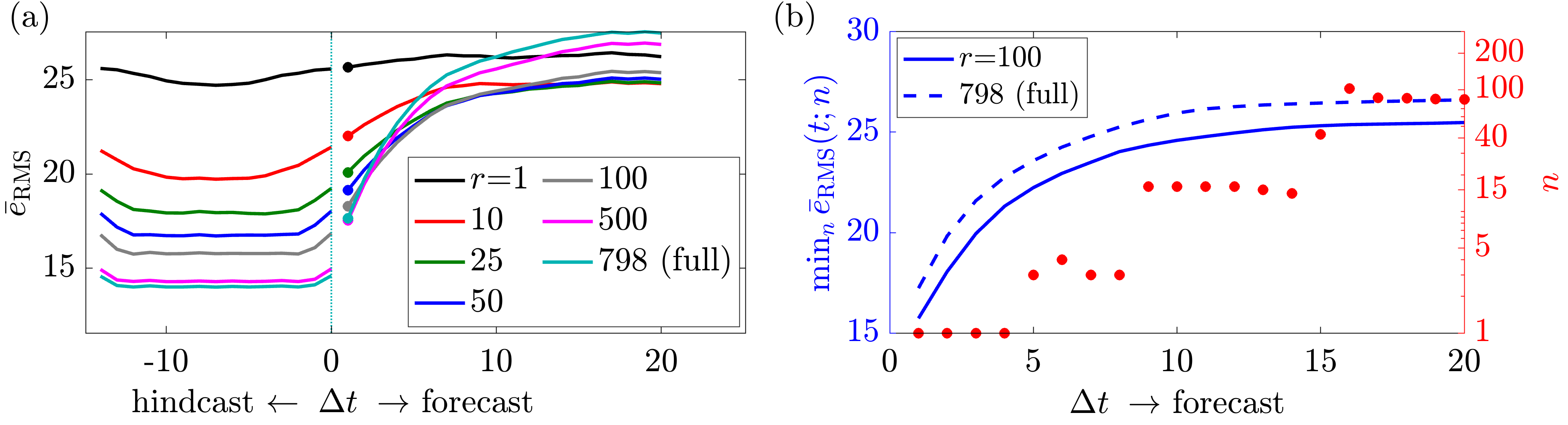}
 \caption{Study of the influence of basis truncation on prediction error for the cavity flow: (a) mean prediction error for different numbers $r$ of retained modes; (b) minimum mean error as a function of forecast time and the corresponding optimal hindcast length. The hindcast and forecast horizons are set to their baseline values of $\Nhind=15$ and $\Nfore=20$, respectively.}
\label{fig:error_rankStudy}
\end{figure}
We next investigate how basis truncation affects forecast accuracy by varying the number of retained modes, $\Nmodes\leq\Nreal$. For the fixed hindcast and forecast horizons shown in Figure \ref{fig:error_rankStudy}(a), the hindcast error decreases as more modes are retained—with the best reconstruction achieved at full rank. In contrast, while retaining a larger number of modes improves accuracy for short forecast times, it actually degrades forecast accuracy at longer times (notably, the rank-1 reconstruction outperforms the full-rank reconstruction for forecast times greater than $10\Delta t$). For the current example, $\Nmodes=100$ represents a good compromise between short- and long-term forecast accuracy. Panel \ref{fig:error_rankStudy}(b) shows the minimum mean RMS error---obtained in the same manner as in Figure \ref{fig:error_25skip_1to200}(b) by selecting the optimal hindcast horizon $\Nhind$---but computed for a reduced rank of $\Nmodes=100$. For comparison, the full-rank optimum from Figure \ref{fig:error_25skip_1to200}(b) is overlaid as a dashed line. It is evident that reducing the rank to 100 modes improves forecast accuracy across the entire forecast horizon. Collectively, the results of the parameter studies in Figures \ref{fig:error_25skip_1to200} and \ref{fig:error_rankStudy} justify our baseline choices for the hindcast horizon and rank reduction parameters, but also highlight that including too many unconverged modes can degrade long-term forecast accuracy by amplifying noise and overfitting transient features.

\begin{figure}[ht!]
 \centering
 \includegraphics[width=1\textwidth,trim={0 2cm 0 0},clip]{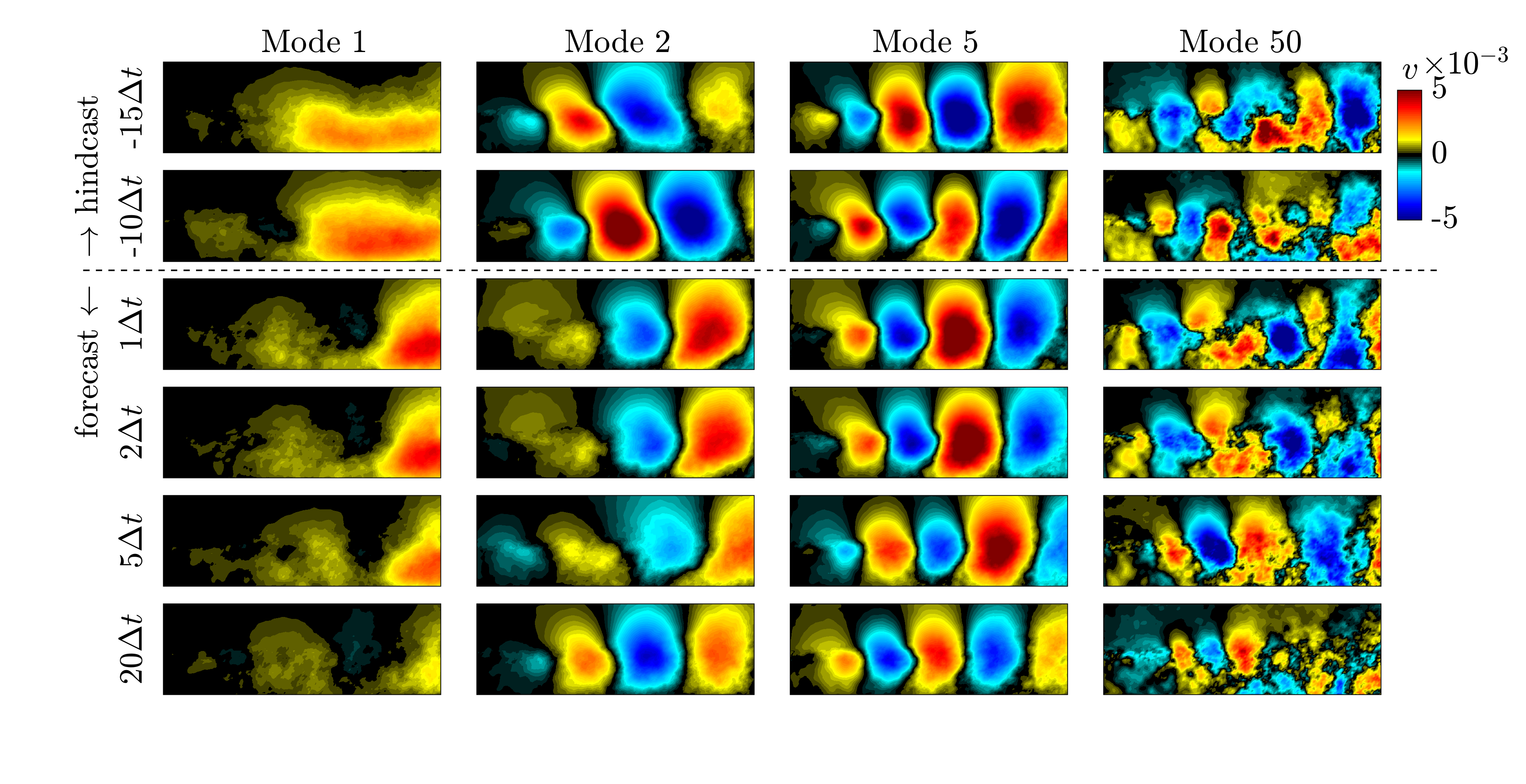}
 \caption{STP modes of the cavity flow for a hindcast horizon of $\Nhind=15$ and a forecast horizon of $\Nfore=20$. A prediction using this basis is shown in Figure \ref{fig:prediction}. The domain is the same as in Figure \ref{fig:flow}.}
\label{fig:modes}
\end{figure}
Figure \ref{fig:modes} shows the time evolution of the $v'$-velocity component of four representative STP modes at several time instances spanning both the hindcast and forecast horizons. The first mode represents a slow modulation of the mean field, with its strongest contributions near the back wall and the bottom of the cavity. Its spatial structure suggests a link to centrifugal modes---a well-known low-frequency recirculation phenomenon in open cavity flows \cite{bres2008instability}. Modes 2 and 5 capture the dominant spatiotemporal structures in the cavity, corresponding to the Rossiter modes highlighted in Figure \ref{fig:flow}(c,d). As hydrodynamic instability waves, these structures are highly coherent, energetic, and oscillatory, which explains their prominence in the STP modes and their contribution to forecasting. In contrast, mode 50, representative of higher-order modes, still displays a discernible structure but appears much noisier. As discussed in Figure \ref{fig:error_rankStudy}, including even higher modes, which tend to be increasingly less converged and noisy, can be detrimental to forecast quality; conversely, truncating these modes can help facilitate noise rejection.

\begin{figure}[ht!]
 \centering
 \includegraphics[width=1\textwidth,trim={0 1.5cm 0 0},clip]{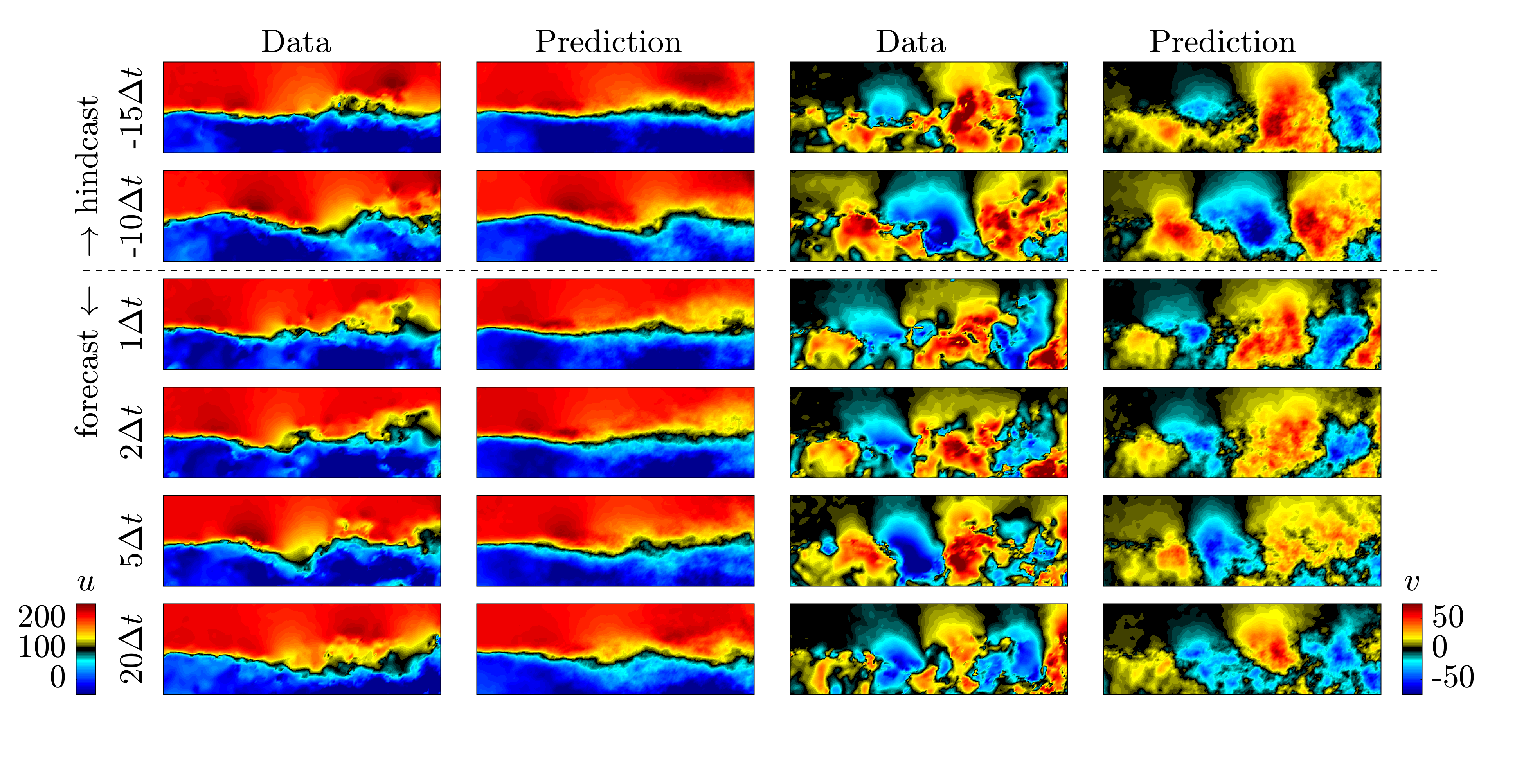}
 \caption{
 Prediction of the cavity flow using a hindcast horizon of $\Nhind=15$, a forecast horizon of $\Nfore=20$, and $\Nmodes=100$ modes retained. The first episode of the test dataset is shown. The domain is the same as in Figure \ref{fig:flow}.}
\label{fig:prediction}
\end{figure}

Figure \ref{fig:prediction} presents the forecast for a single episode, the first one in the test dataset, using the baseline parameters, $\Nhind=15$, $\Nfore=20$, and $\Nmodes=100$. The ground truth (i.e., the testing data) is compared to the forecast at the same time instances as in Figure \ref{fig:modes}. To facilitate interpretation, the temporal mean fields have been added back to the velocity fields. At the two representative time steps within the hindcast horizon, the forecast closely resembles the data, at least up to the level of detail the truncated modal basis computed from the training data is able to capture. Over the forecast horizon, the correspondence is less precise; however, particularly for the $v$-component, which accentuates large-scale coherent patterns, the spatiotemporal evolution of the instability waves convecting along the cavity can still be tracked. Even at the final forecast time step, $20\Delta t$, the dominant flow structure visible in the data---a wave spanning approximately three wavelengths across the field of view---is discernible in the forecast with the correct phase and approximate amplitude. While the turbulent and chaotic nature of the flow inherently limits predictability, the forecast's ability to capture the evolution of these large-scale structures is noteworthy.

\subsection{Performance}\label{sec:performance}
We study two aspects of the method’s performance. First, we examine how forecast accuracy depends on the amount of available training data, evaluating this sensitivity for both datasets. Second, we compare the STP forecast error with that of Long Short-Term Memory (LSTM) regression neural networks. This comparison is performed only for the stationary cavity data, where the same set of spatial-only POD modes can be used for both hindcasting and forecasting. In contrast, the transient nature of the supernova data would require a different dimensionality-reduction strategy for a simple LSTM model to be applicable, which is beyond the scope of this work.

\begin{figure}[ht!]
 \centering
 \includegraphics[width=1\textwidth,trim={0 0cm 0 0},clip]{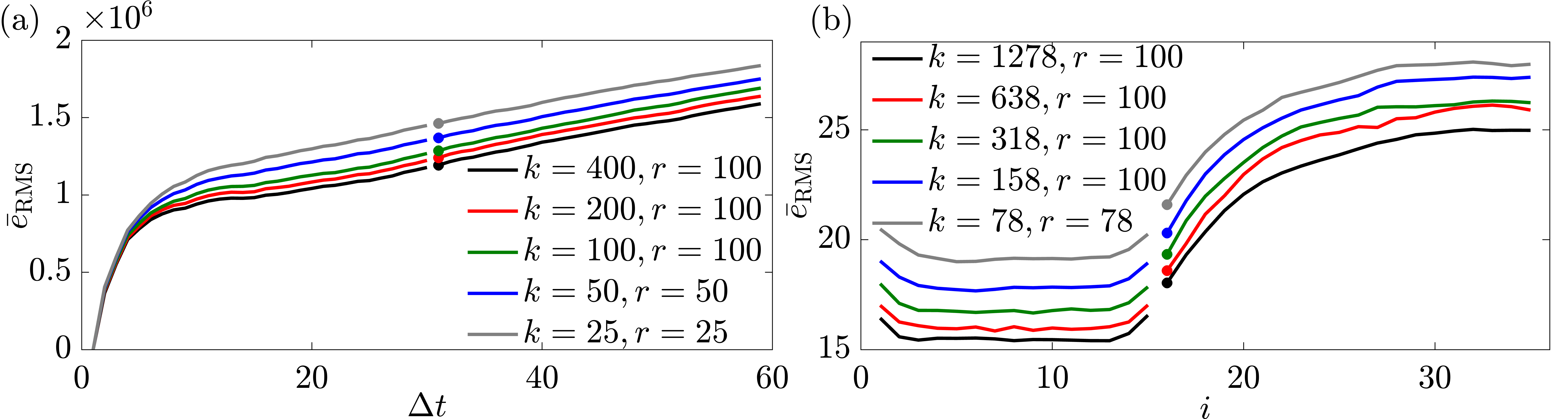}
 \caption{Sensitivity to training ensemble size: (a)  supernova explosion; (b) cavity flow. The hindcast and forecast horizons are fixed, while the training ensemble size is systematically reduced by halving the available training data. The truncation rank is set to $\Nmodes=100$; when $\Nreal\leq100$, full-rank prediction with $\Nmodes=\Nreal$ is employed.}
\label{fig:kStudy}
\end{figure}

In the study of sensitivity to training ensemble size presented in Figure \ref{fig:kStudy}, we set the hindcast and forecast horizons to $\Nhind=15$ and $\Nfore=20$, respectively. The truncation rank is fixed at $\Nmodes=100$, except when the ensemble size $\Nreal$ is smaller than 100, in which case the prediction is performed at full rank, $\Nmodes=\Nreal$. Similar trends are observed in both datasets: as expected, the reconstruction accuracy decreases with fewer training episodes, yet the relative hindcast accuracy remains a reliable qualitative indicator of forecast accuracy. Notably, important trends, such as the practical forecasting horizon for the cavity flow being limited to about 10–15$\Delta t$, are consistently observed even with the smallest amount of training data used, namely 78 episodes for the cavity flow and 25 episodes for the supernova case. For both datasets, each halving of the training ensemble size results in a roughly equal incremental increase in prediction error. The relatively larger errors observed for the smallest ensemble size, particularly pronounced for the supernova data in Figure \ref{fig:kStudy}(a), can be attributed to simultaneous rank reduction necessitated by fewer available training episodes. The consistent increase in prediction error with reduced training data strongly suggests that the ability of the STP basis to represent the data is directly linked to the statistical convergence of the sample covariance matrix. Since the eigenvalues measure the variance (or energy) captured by each mode, constructing a robust and converged basis requires a sufficiently large ensemble to accurately estimate these eigenvalues and their corresponding modes. With fewer training samples, the basis estimation becomes less reliable, systematically degrading forecast accuracy. In summary, this study underscores the importance of using sufficiently large training ensembles for accurate forecasting, but also highlights the robustness of the STP method even when data availability is limited.

Next, we compare the STP forecast error to that obtained from standard LSTM regression neural networks. We specifically use MATLAB's standard LSTM implementation \cite{MATLAB_LSTM2024}, as it is widely accessible, well-documented, and reproducible. As direct prediction in physical space is computationally prohibitive, we instead perform predictions on POD coefficients, that is, we use space-only POD for dimensionality reduction. For the stationary cavity flow dataset, the space-only POD modes are computed using the full training dataset, represented by the $\Nx \times K$ snapshot matrix
$\vb{U} = \qty[ \vb{u}_1 \; \vb{u}_2 \; \cdots \; \vb{u}_K]$
where $K=12800$ is the total number of training snapshots, differing from the number of training episodes, $\Nreal=1278$, due to the overlapping segmentation described in Section \ref{sec:cavity}. As before, the state vector includes both velocity components. The LSTM network is trained on the corresponding temporal POD coefficients as latent variables, with rank truncation levels set to $\Nmodes=25, 100,$ and $500$. These levels capture approximately 68\%, 83\%, and 93\% of the total variance, respectively, compared to 71\% for $\Nmodes=100$ hindcast modes. This should, in principle, guarantee that the LSTM netowrk is not disadvantaged by the choice of rank truncation. The LSTM configuration includes a sequence input layer, one LSTM layer with 128 hidden units as default, also varied to 64 and 256 units, and a fully connected output layer. Training targets are generated by shifting the POD coefficients by one time step to forecast future values, and data are normalized to zero mean and unit variance, following standard practices \cite{MATLAB_LSTM2024}. The network is trained using the Adam optimizer with a constant learning rate of 0.001 for 200 epochs, with data shuffled each epoch. With the default mini-batch size of 128, the training data is divided into 9 mini-batches per epoch, resulting in a total of 1800 iterations.

\begin{figure}[ht!]
 \centering
 \includegraphics[width=1\textwidth,trim={0 0cm 0 0},clip]{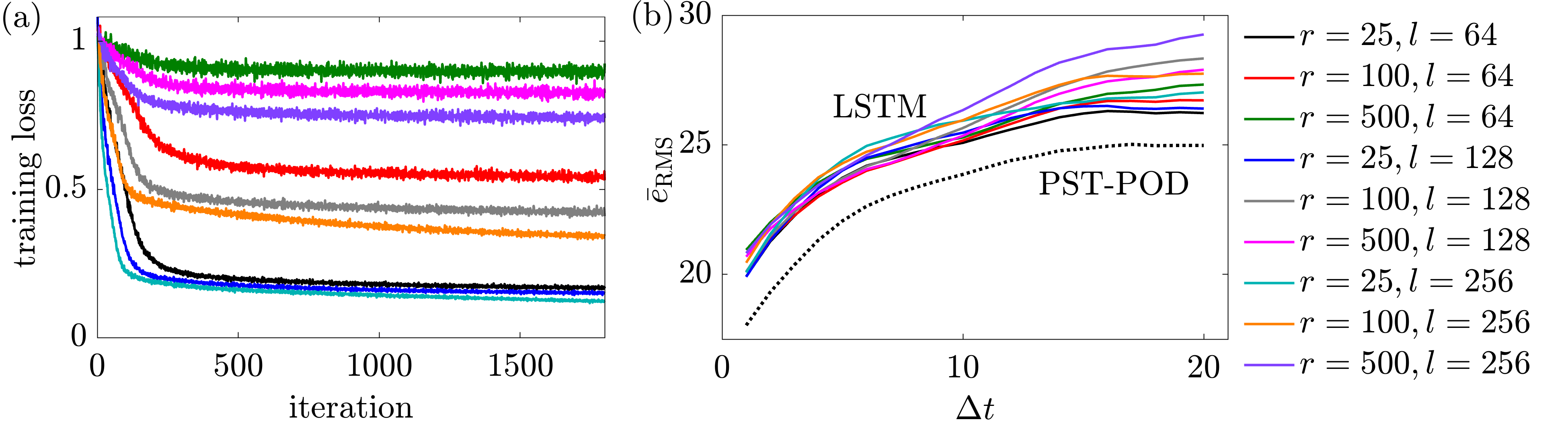}
 \caption{Comparison of STP and standard LSTM regression networks with varying numbers of hidden units ($l$) and learned POD time coefficients ($\Nmodes$): (a) training loss; (b) mean prediction error. Both methods use identical training data of $\Nreal=1278$ episodes and the same hindcast horizon. The STP prediction with $\Nmodes=100$ from Figure \ref{fig:spectrum_and_error} is reproduced for comparison.}
\label{fig:cavity_LSTM}
\end{figure}
Figure \ref{fig:cavity_LSTM} summarizes the training loss convergence and mean prediction error, comparing STP to the LSTM results across various ranks and hidden unit configurations. We also tested hindcast horizons of $\Nhind=5$ and 30, and deeper network architectures with three LSTM layers, but observed no significant qualitative differences. The STP prediction using the parameters identified in Section \ref{sec:cavity} is included as a reference. Two primary trends can be observed in Figure \ref{fig:cavity_LSTM}(a). First, convergence of the training loss is strongly influenced by the number of POD coefficients used as inputs to the LSTM: the best convergence is observed for the smallest number of modes, $\Nmodes=25$. A secondary, less pronounced trend is that increasing the number of hidden units, $l$, generally improves convergence for a fixed number of coefficients. Examining the mean prediction error in Figure \ref{fig:cavity_LSTM}(b), the LSTM configuration with the smallest number of POD coefficients ($\Nmodes=25$) and fewest hidden units ($l=64$) performs best for forecast horizons longer than $10\Delta t$, whereas networks with larger numbers of units and coefficients yield slightly lower errors at shorter times, though these differences are modest. A plausible explanation is that simpler models (with fewer parameters and lower-rank representation) generalize better at longer forecast horizons, while more complex networks tend to slightly outperform at shorter forecast horizons. Crucially, the STP method consistently achieves lower prediction errors than the tested LSTM architectures across the entire forecast horizon. We emphasize, however, that our comparison was conducted using a standard LSTM architecture and the method-of-snapshots POD as a straightforward dimensionality-reduction strategy. The motivation for this choice was to ensure fairness in computational costs and reproducibility. While these results clearly demonstrate the capability and effectiveness of STP relative to a representative, widely recognized state-of-the-art method, we acknowledge that alternative neural network architectures or dimensionality-reduction strategies---beyond the scope of the current study---may likely yield more competative results.

The primary computational burden of our approach lies in constructing the hindcast correlation matrix and performing its eigenvalue decomposition, or alternatively, the singular value decomposition of the hindcast data matrix. Although this can be challenging---particularly in terms of memory usage---for very large datasets, streaming algorithms that are well-established \cite{brand2002incremental,SchmidtTowne_2018_CPC} can mitigate this burden. To put this into perspective, however, all computations reported in this article were performed using full matrix decompositions of the complete databases (1.1 GB for the cavity flow and 24.9 GB for the supernova), and were completed within minutes for individual computations (excluding parameter sweeps) on a laptop with 64 GB of memory and an Apple M1 Max chip. The compute times of the STP predictions and the LSTM neural networks were comparable.

\section{Discussion}\label{sec:discussion}

We present a method for data-driven forecasting of both transient and stationary phenomena based on an ensemble of realizations of the transient process. Originating from the theory of POD, our method leverages correlation along with the inherent properties of POD---namely, orthogonality and optimality---to generate predictions. For a given ensemble of realizations and a fixed prediction horizon, the only tunable parameters are the rank truncation and the length of the hindcast horizon; no additional hyperparameters are required. In particular, the forecast horizon does not affect the prediction. The inherent rank reduction and latent space compression are key features of the method, and we contend that the conceptual and algorithmic simplicity of the approach is a major strength.

For both datasets used to test the method, forecasts are produced that appear as continuous extensions of the hindcast horizon into the future. Evaluation on the test data via the conventional 80\%/20\% split revealed that the forecasts effectively capture the evolution of the dominant spatiotemporal patterns in each dataset. The modes, which form a hierarchical basis capturing the variability of the trajectories, exhibit increasing geometric complexity. In the cavity flow case, these modes are clearly linked to physical phenomena that account for their persistence in the data. Furthermore, the hindcast serves as a reliable indicator of forecast accuracy at short prediction times and acts as a lower bound for the forecast error. Notably, only the cavity flow case benefited from rank truncation. This is likely due to the overlapping segmentation used to construct the training episodes and the resulting incorporation of redundant information, which aids the convergence of the leading modes but also inflates the basis with additional, unconverged modes that capture little variance. In contrast, for the supernova data, the full-rank prediction was most accurate. The likely reason is the immense complexity of the expanding shell structures in the individual trajectories and the limited availability of training data to capture the highly anisotropic dynamics characterized by localized structural changes. In this case, the initial expansion stages, emerging from randomly seeded multiple nuclei, determine the subsequent dynamics, and our method successfully predicts the expansion shell pattern over extended forecast horizons.

For a quantitative comparison with state-of-the-art forecasting methods, we chose a LSTM regression neural network to predict POD coefficients of the cavity flow example. To ensure fairness, we confirmed that the two different POD bases---STP and classical method-of-snapshots POD—capture comparable amounts of energy. Despite systematically varying key hyperparameters, the LSTM models did not yield more accurate forecasts than STP. While this does not preclude the possibility that other conventional or machine learning algorithms---or LSTM combined with a different dimensionality-reduction strategy---could outperform it, the results nonetheless underscore the robustness and effectiveness of the proposed method, establishing STP as a competitive benchmark for future forecasting studies.

While in our examples we predict future states at the same spatial locations and for the same state variables, the approach is not inherently limited to this configuration. The spatial domains of the hindcast and forecast modes can differ, and the variables used for hindcasting need not be identical to those being forecasted. This flexibility opens up applications such as forecasting from sensor data, where the hindcast may differ from the forecast both in terms of the variables of interest and locations. Other promising directions include parametric extensions of the method to account for system variations across different physical or geometric parameters \cite{barrault2004empirical}, as well as conditional sampling strategies aimed at predicting imminent extreme events based on flow state precursors \cite{schmidt2019conditional}.

\clearpage

\paragraph{Acknowledgements} Many thanks to Brandon Yeung for suggesting the supernova dataset and providing the MATLAB I/O script, and to Yang Zhang and Lou Cattafesta for generously sharing the cavity flow data. Support from the Air Force Office of Scientific Research (grant FA9550-22-1-0541), the National Science Foundation (grant CBET-2046311), and the Office of Naval Research (grant N00014-23-1-2457) is gratefully acknowledged.

\bibliographystyle{alpha}
\bibliography{sample}

\end{document}